\documentclass{sig-alternate}
\usepackage{flushend}
\usepackage{graphicx}
\usepackage{setspace}

\newenvironment{mybox}{\begin{tabular}{@{}l@{}}}{\end{tabular}}
\newcommand{\etal}{et~al.~}

\begin{document}
%
% --- Author Metadata here ---
%\conferenceinfo{KDD}{'14 New York, NY USA}
%\CopyrightYear{2007} % Allows default copyright year (20XX) to be over-ridden - IF NEED BE.
%\crdata{0-12345-67-8/90/01}  % Allows default copyright data (0-89791-88-6/97/05) to be over-ridden - IF NEED BE.
% --- End of Author Metadata ---

\title{Efficient multivariate kernels for sequence classification}
%\subtitle{Efficient multivariate kernels for sequence classification}

\numberofauthors{1} %  in this sample file, there are a *total*
\author{
% 1st. author
\alignauthor
Pavel P. Kuksa\\
%       \affaddr{NEC Laboratories America Inc}\\
%       \affaddr{4 Independence Way}\\
%       \affaddr{Princeton, NJ}\\
       \email{pavel@pkuksa.org}
}

\maketitle

\begin{abstract} \small\baselineskip=9pt

Kernel-based approaches for sequence classification
have been successfully applied to a variety of domains,
including the text categorization, image classification, speech analysis, 
biological sequence analysis, time series and music classification, where
they show some of the most accurate results.

Typical kernel functions for sequences in these domains
(e.g., bag-of-words, mismatch, or subsequence kernels)
are restricted to {\em discrete univariate} (i.e. one-dimensional)
string data, such as sequences of words in the text analysis,
codeword sequences in the image analysis, or nucleotide or amino acid
sequences in the DNA and protein sequence analysis.
However, original sequence data are often of real-valued multivariate nature, i.e. are not univariate and discrete as required by typical
$k$-mer based sequence kernel functions.

In this work, we consider the problem of the {\em multivariate}
sequence classification (e.g., classification of multivariate music sequences, or
multidimensional protein sequence representations).
To this end, we extend {\em univariate} kernel functions typically used in
sequence domains and propose efficient
{\em multivariate} similarity kernel method
(MVDFQ-SK) based on  (1) a direct feature quantization (DFQ) of
each sequence dimension in the original {\em real-valued}
multivariate sequences and
(2) applying novel multivariate discrete kernel measures on these multivariate discrete DFQ sequence representations to more accurately capture
similarity relationships among sequences and improve classification
performance.

Experiments using the proposed MVDFQ-SK kernel method show excellent
classification performance on three challenging music classification tasks
as well as protein sequence classification with significant 25-40\%
improvements over univariate kernel methods and existing
state-of-the-art sequence classification methods.

\end{abstract} 

%A category including the fourth, optional field follows...
%\category{D.2.8}{Software Engineering}{Metrics}[complexity measures, performance measures]


\section{Introduction}
\label{Sec:Intro}
Large-scale sequence analysis has become
an important task in data mining inspired by numerous applications such as
the document and text classification or the analysis of time series, music data, or biological sequences.  
Classification of string data, i.e. univariate sequences of
{\em discrete} symbols (such as words, amino acids, codewords),
has attracted particular attention and has led to a number of
new algorithms~\cite{PierreBaldiFoldRecogn,PROFILE-K,MISMATCH-K,icpr2010sssk,LargeScaleGenomicSeqSVM,smkcvpr,binetcauchy2007,vishwanathan,gldseqkernel2002,plsa2009music}.
%These algorithms often exhibit state-of-the-art performance on tasks such as structural and functional protein sequence classification,
%time series classification and clustering, music genre recognition and document topic elucidation.

In particular, kernel-based approaches for sequence classification
show some of the most accurate results for a variety of
problems, such as
text categorization~\cite{icpr2010sssk,multinomial2005text},
image classification~\cite{smkcvpr},
speech analysis~\cite{gldseqkernel2002},
biological sequence analysis~\cite{LargeScaleGenomicSeqSVM,icpr2010sssk,nips2008inexact},
time series and music classification~\cite{binetcauchy2007,icpr2010sssk}.

In a kernel-based framework, the classification of the sequence
$X=x_1,\ldots,x_{n_X}$
is based on a kernel function $K(X,Y)$ which is computed
to measure the similarity between pairs of sequences $X$ and $Y$.
For instance, given a set of positive training instances ${\cal C}^{+}$ and
a set of negative training instances ${\cal C}^{-}$, the SVM~\cite{VAPNIKLEARNING}
learns a classification function of the following form
\begin{equation}
f(X) = \sum_{Y^{^+}\in {\cal C}^{^+}} \alpha^{^+}_{Y^{^+}} K(X,Y^{^+}) - \sum_{Y^{^-} \in \cal{C}^{^-}} \alpha^{^-}_{Y^{^-}} K(X,Y^{^-})
\end{equation}

\if 0
In a kernel-based framework for sequence classification,
a family of state-of-the-art kernel functions
$K(X,Y)$
for measuring similarity between pairs of sequences $X$ and $Y$
relies on fixed length, substring spectral feature representations and
the notion of inexact matching (matching with mismatches), c.f.~\cite{PROFILE-K,MISMATCH-K}.
In these approaches, a sequence $X$ is represented as the spectra (histogram of counts) of
all short $k$-long substrings ($k$-mers) contained within the sequence with
possibly up to $m$ mismatches.
Initial work, e.g.,~\cite{MISMATCH-K,shaweTaylorBook,fastInexact},
has demonstrated that this class of $k$-mer-based similarity functions
can be computed using trie-based approaches in $O(k^{m+1}|\Sigma|^m(|X|+|Y|))$, for strings $X$ and $Y$
with symbols from alphabet $\Sigma$ and up to $m$ mismatches
between $k$-mers.
More recently, \cite{nips2008inexact} introduced {\em linear time} $O(c_{k,m}(|X|+|Y|))$ algorithms with alphabet-independent complexity for these
$k$-mer based string kernel functions. 
\fi

Typical $k$-mer-based kernel functions (e.g., mismatch or spectrum kernels,
gapped and  wildcard kernel functions~\cite{fastInexact,LodhiString})
essentially rely on {\em symbolic} {\em Hamming-distance} based matching
of one-dimensional (1D) $k$-mers ($k$-long substrings) in the input sequences.
For example, given one-dimensional (1D) sequences $X$ and $Y$ over alphabet $\Sigma$ (e.g.,
amino acid sequences with $|\Sigma|$=$20$), the {\em spectrum}-$k$
kernel~\cite{SPECTRUM-K} and the {\em mismatch}-($k$,$m$) kernel~\cite{MISMATCH-K} measure similarity between sequences as
\begin{align}
\nonumber K(X,Y|k,m) & = \left < \Phi_{k,m}(X), \Phi_{k,m}(Y) \right > \\
\nonumber & = \sum_{\gamma \in \Sigma^k} \Phi_{k,m}(\gamma|X) \Phi_{k,m}(\gamma|Y)\\
\nonumber & = \sum_{\gamma \in \Sigma^k} \left( \sum_{\alpha \in X} I_m(\alpha,\gamma) \right) \left( \sum_{\beta \in Y} I_m(\beta,\gamma) \right) \\ 
 	& = \sum_{\alpha \in X} \sum_{\beta \in Y} \sum_{\gamma \in \Sigma^k} I_m(\alpha,\gamma) I_m(\beta,\gamma)
\label{Eq:MK}
\end{align}
where
\begin{equation}
   \Phi_{k,m}(\gamma|X)  =  \left( \sum_{\alpha \in X} I_m(\alpha,\gamma)
                 \right)
\end{equation}
is the number of occurrences (possibly with up to $m$ mismatches)
of the $k$-mer $\gamma$ in $X$,
and 
the matching/indicator function $I_m(\alpha,\gamma)=1$ if $\alpha$ is in the mutational neighborhood $N_{k,m}(\gamma)$ of $\gamma$, i.e. $\alpha$ and $\gamma$ are at the Hamming distance of at most $m$.
This kernel (Eq.~\ref{Eq:MK}) essentially amounts to
a cumulative Hamming-distance based pairwise comparison of all $k$-mers
$\alpha$ and $\beta$ contained in sequences $X$ and $Y$, respectively, with
maximum number of mismatches $m$.
The level of similarity of each pair of substrings $(\alpha,\beta)$ here
is indicated by the number of identical substrings in the mutational neighborhoods $N_{k,m}(\alpha)$ and $N_{k,m}(\beta)$ of $\alpha$ and $\beta$,
$\sum_{\gamma \in \Sigma^k} I_m(\alpha,\gamma) I_m(\beta,\gamma)$ (Eq.~\ref{Eq:MK}).
For the spectrum kernel ($m$=0), this similarity level is simply the exact matching of $\alpha$ and $\beta$.

On the other hand, in many practical applications
input sequences are {\em multivariate}, i.e. input data is in the form
of sequences of {\em $R$-dimensional} real-valued feature vectors, as opposed to one-dimensional discrete strings.
This is the case, for instance, in commonly used
MFCC representations for music data as series of $13$-dimensional
MFCC feature vectors
(e.g.,~\cite{tzan2002,sigir2003music,spectralcepstral2009}) extracted
from short time segments, or 20-dimensional profile representations
of protein sequences as series of probabilistic amino acid substitution
vectors in biological sequence analysis~\cite{PROFILE-K,gribskov87pnas}.

Such original, multivariate real-valued feature sequences are
typically transformed into {\em univariate sequences}
in order to apply a univariate string kernel method such as
spectrum or mismatch, e.g.,~\cite{smkcvpr,plsa2009music,icpr2010sssk,nips2008inexact}. For example, this transformation is frequently
accomplished by applying
a vector quantization (VQ) algorithm to feature vectors thus transforming
a multivariate $R$-dim real-valued sequence into a discrete univariate {\em codeword} sequence.

In contrast,
in this work we consider an alternative approach to real-valued
multivariate sequence classification which directly
exploits these richer multivariate ($R$-dimensional) sequences
(e.g., MFCC feature sequences for music,
or sequences of physico-chemical amino acid descriptors for proteins).
In this approach, the $R$-dimensional multivariate sequences are considered as
$R \times |X|$ feature - spatio/temporal matrices 
with rows corresponding to feature dimensions and columns corresponding to
temporal or spatial coordinates.
Using these representations, we propose an
efficient discrete multivariate kernel method (MVDFQ-SK) based on (1)
a direct feature quantization (DFQ) (Sec.~\ref{Sec:dfq})
of the original multivariate sequence,
and (2) novel manifold-based discrete multivariate kernel 
functions applied to these discrete DFQ representations
(Sec.~\ref{Sec:MVDFQSK},~\ref{Sec:manifold}). 
%that allows {\em linear time} inexact matching and classification of
%sequence inputs in the form of sequences of $R$-dimensional feature vectors
%(Sec.~\ref{Sec:2DSK}). 
The developed approach
is applicable to %modeling of sequences in
a wide range of
sequence domains, both {\em discrete}- and {\em real}- valued,
such as {\em music}, {\em images}, or {\em biological sequences}.

Experiments using the new multivariate direct feature quantization kernels 
(MVDFQ-SK) kernels on music genre and artist
recognition, as well as protein sequence classification tasks show excellent predictive performance (Sec.~\ref{Sec:exp})  with significant
{\em 25\%-40\%} improvements in predictive accuracy over univariate kernel functions and a number of other state-of-the-art sequence classification
methods.

\section{Related work}
%Over the past decade, 
Recently, a large variety of methods have been proposed to solve the sequence classification problem, including {\em generative}, such as HMMs, or {\em discriminative} approaches.
Among the discriminative approaches, string kernel-based~\cite{VAPNIKLEARNING,shaweTaylorBook} methods provide some of the most accurate results~\cite{PROFILE-K,MISMATCH-K,LargeScaleGenomicSeqSVM,fastInexact,icpr2010sssk,rationalKernels,smkcvpr,multinomial2005text}
in many sequence analysis tasks. 
\if 0
The key idea of basic string kernel methods is to first
map sequences of variable length
into a fixed-dimensional vector space
using a feature map $\Phi(\cdot)$.
Then, in this space a standard
classifier such as a support vector machine (SVM)~\cite{VAPNIKLEARNING} can then be applied.  As SVMs require only inner products
between examples in the feature space, rather than the feature vectors themselves, 
%one is only required to provide the {\em string kernel}:
one can define a {\em string kernel} $K(\cdot,\cdot)$
which computes the inner product in the feature space without explicitly computing the feature vectors:
\begin{equation}
K(X,Y) = \langle \Phi(X), \Phi(Y) \rangle,
\end{equation}
where $X,Y \in D$, $D$
 is the set of all sequences composed of elements which take on a finite set of possible values from the alphabet $\Sigma$.
%.g., sequences of words in our case,  and $\phi: S \to \R^m$ is a feature
%mapping from a word sequence (text) to a $m$-dim. feature vector.
\fi

In the kernel-based approaches, the similarity
between sequences $X$ and $Y$ is frequently computed based on
the co-occurrence of string features (e.g., $k$-mers),
as in spectrum kernels~\cite{SPECTRUM-K} or
substring kernels~\cite{vishwanathan}.
Inexact comparison of the sequences in this framework is typically achieved using different families of mismatch~\cite{MISMATCH-K} or 
profile~\cite{PROFILE-K} kernels.
Both spectrum-$k$ and mismatch-($k$,$m$) kernels directly extract string
features ($k$-mers) from the observed sequence, $X$. On the other hand, the 
profile kernel, proposed by Kuang \etal in~\cite{PROFILE-K}, first builds 
a $20\times|X|$-dim profile~\cite{gribskov87pnas} $P_X$ and then derives
a similar  $|\Sigma|^k$-dimensional representation from $P_X$.
Such profile representations have been shown to perform well
in protein sequence analysis~\cite{PROFILE-K,JMLRProfileMultiClass}.
Constructing the profile for each sequence may not be practical in
some application domains, since the size of the profile is dependent on the
size of the alphabet set, as well as the difficulty of defining a
general sequence similarity search algorithm (e.g., as PSI-BLAST) for
non-biological sequence domains.
While for bio-sequences $|\Sigma|=4$ or $20$, 
for music or text classification $|\Sigma|$ can potentially be very large, 
on the order of tens of thousands of symbols.

The existing string kernel methods essentially amount to the analysis of
univariate (i.e. one-dimensional) sequences over finite alphabets
$\Sigma$ with one-dimensional $k$-mers as basic sequence features.
However, original input sequences are often in the form of
{\em sequences of feature vectors}, i.e.
each input sequence $X$ is a {\em sequence of identically sized ($R$-dim) feature vectors}
which
could be considered as a $R \times |X|$ feature matrix.
%(i.e. multivariate or $2D$ sequence).

Examples of these multivariate feature sequences include 
\begin{itemize}
\itemsep -1pt
\topsep -1pt
\partopsep -1pt
\item
{\em Music data}. A music sequence $X$ in the commonly used MFCC feature
representation~\cite{spectralcepstral2009,sigir2003music} is a sequence of 13-dimensional MFCC feature vectors, i.e. a multivariate sequence of size $13\times|X|$.
%is typically a sequence of 13-dim. MFCC features, i.e. 2D sequence of size $13\times|X|$.
\item
{\em Image data}. An image can be considered as a
multivariate sequence of feature vectors
extracted from image patches
%extracted from
%image patches obtained by decomposition of the image
%into a regular grid of smaller image blocks
(e.g., as in~\cite{smkcvpr}); 
\item
{\em Biological data}. Protein sequences
can be viewed as {\em profiles}~\cite{PROFILE-K},
or as multivariate sequences of $R$-dim feature vectors
describing physical/chemical properties of individual amino acids~\cite{bmc2010phychem}.
\end{itemize}

While typical string kernel methods essentially use {\em symbolic Hamming
distance-based} matching (as in Eq.~\ref{Eq:MK}), 
recent work in~\cite{sdm2012distsk} introduced the so-called generalized similarity ({\em non-Hamming}) kernels that allow to incorporate general similarity metrics ${{\cal S}(\cdot,\cdot)}$ into similarity evaluation and improve performance compared
to symbolic Hamming distance-based matching~\cite{sdm2012distsk}.
In particular, most related to the current work, are the distance-preserving symbolic embedding kernels~\cite{sdm2012distsk} which use similarity
hashing~\cite{spectralHashing} to obtain {\em binary} representations for sequences such that
Hamming distance $h(\cdot,\cdot)$ between these binary representations is proportional
to the original similarity score ${{\cal S}(\cdot,\cdot)}$~\cite{sdm2012distsk}.
In contrast, the direct feature quantization (DFQ) method proposed in this work
results in more accurate {\em non-binary} representations that are simpler as they do not require Hamming embedding learning step as in~\cite{sdm2012distsk}, and display higher accuracy (see Experiments, Sec.~\ref{Sec:exp})
compared to the binary Hamming-based distance-preserving embedding.

Related methods for the time series classification have also been introduced and include a large variety of methods, e.g., kernels on dynamical systems~\cite{binetcauchy2007}, or alignment-based methods~\cite{cuturi07kernel,dtak2002} with a quadratic time complexity.
We will compare in the experiments with a number of these 
%typically used
methods for time series. % (e.g., multivariate autoregressive models).

In this work, in contrast to kernel methods on univariate string representations, we aim at methods that {\em directly} exploit 
multivariate sequence representations to improve accuracy and propose a family
of efficient, linear-time {\em discrete multivariate} similarity kernels
(MVDFQ-SK, MVDFQM-SK) using direct feature quantization (DFQ) and manifold kernel embedding (Sec.~\ref{Sec:dfq},~\ref{Sec:manifold}).
We show empirically (Sec.~\ref{Sec:exp}) that proposed MVDFQ/MVDFQM kernels
and manifold embedding (Sec.~\ref{Sec:manifold}) provide effective
improvements in practice over traditional univariate (1D) VQ-based
sequence kernels, binary similarity-preserving embedding
kernels~\cite{sdm2012distsk}, as well as other state-of-the-art sequence classification methods for a
number of challenging classification problems.

\section{Multivariate Direct Feature Quantization Method}
\label{Sec:MVDFQmethod}
In a typical sequence classification setting, string kernels are restricted to the univariate
string data, e.g., word sequences in text analysis, amino acid sequences
in the biological sequence analysis, or codeword sequences in the time series
analysis~\cite{smkcvpr,LodhiString,icpr2010sssk,MISMATCH-K}.

\begin{figure}[!ht]
  \centering
  \includegraphics[width=0.30\textwidth]{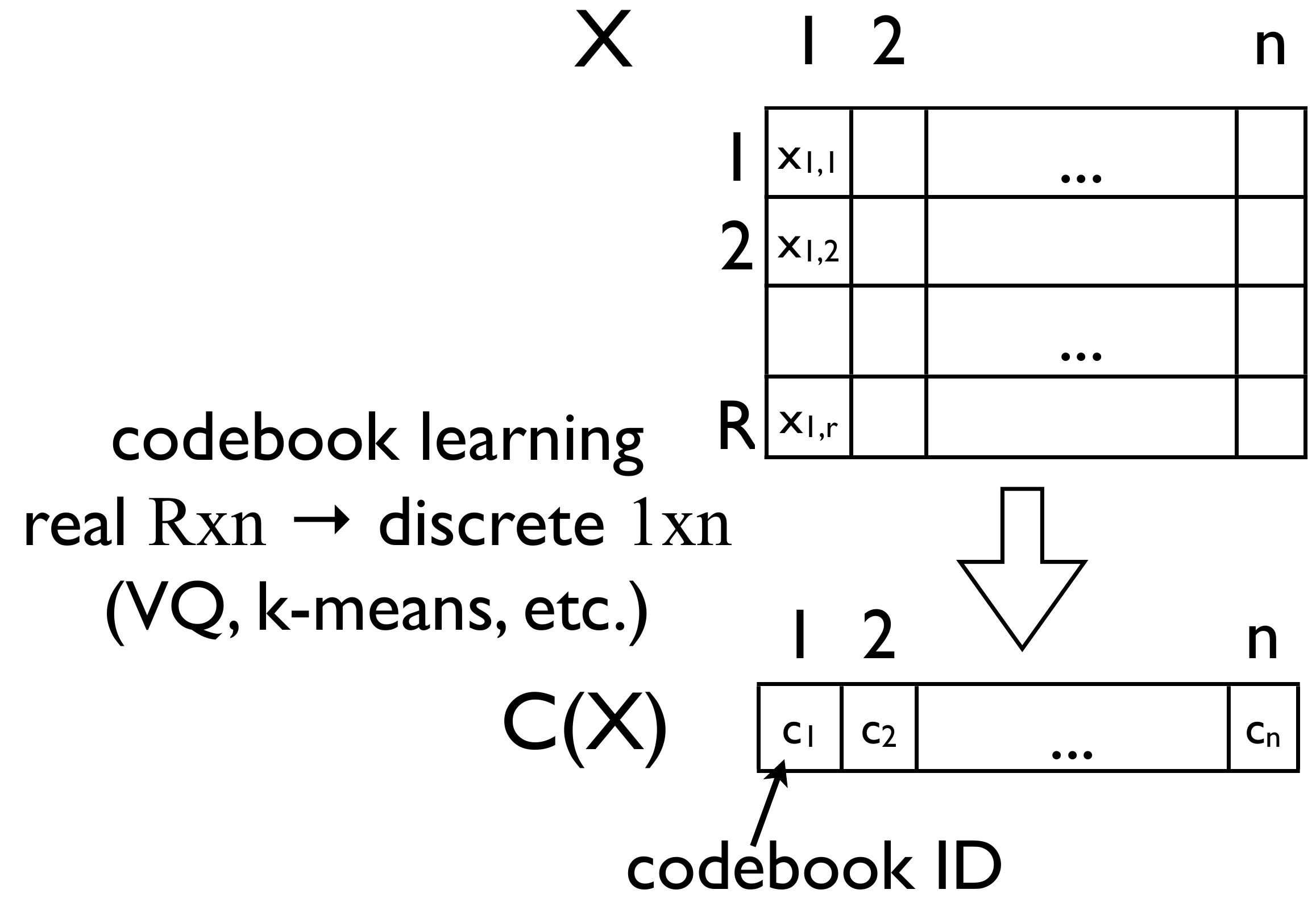}            
  \caption{\small{A typical codebook-based representation for a $R$-dim sequence $X$: $R$-dimensional input feature vectors are encoded using corresponding codebook IDs. Univariate string kernels are used to compute similarity between sequences (e.g.,~\cite{smkcvpr,icpr2010sssk})}}
  \label{Fig:vq_2d_1d}
\end{figure}

In order to apply these univariate kernel functions to
multivariate ($R$-dimensional) sequences, individual feature vectors
at each position in the sequence
in the widely used {\em codebook learning framework} are
first encoded using codebook IDs (Figure~\ref{Fig:vq_2d_1d}), then 
%In a typically used {\em codebook learning} framework for classification, input multivariate sequences (or sets) of $R$-dim. features vectors are typically first encoded using
%codebook IDs (Figure~\ref{Fig:vq_2d_1d}), then
standard univariate
string kernel methods can be applied on these discrete
codeword sequence representations (see e.g.,~\cite{smkcvpr,icpr2010sssk}).

As illustrated in the Fig.~\ref{Fig:vq_2d_1d}, the $R$-dim features vectors from
input sequences are first quantized (clustered) to obtain a codebook ${\cal C}$,
a set of codebook (prototype) vectors, ${\cal C} = \{C_1,C_2,...,C_N\}$,
for instance, by applying a Vector Quantization (VQ) algorithm. 
Then a multivariate input sequence $X$, a sequence of $n=|X|$ identically
sized (real-valued) $R$-dimensional feature
vectors,
\[ {\mathbf X}=(x_1,x_2,\ldots,x_n), x_i \in {\cal R}^{R} \; \forall i
\]
is encoded as a {\em univariate (1D) discrete sequence} $c(X)$ of codebook IDs
\[ {\mathbf c(x)}=(c_1,c_2,\ldots,c_n), c_i \in \{1 \ldots D\} \; \forall i \]
by mapping each of the vectors $x_i$ to the nearest codeword vector $c_i$ in the
codebook ${\cal C}=\{C_1,C_2,\ldots,C_D\}$.
The resulting codeword sequence ${\mathbf c(x)}$ is essentially a discrete sequence over
finite alphabet $\Sigma=\{1,\ldots,D\}$.
Univariate (1D) string kernels can then be used for classification with
SVM.

In contrast to these commonly employed codebook-based
univariate representations, in this work we consider
an alternative {\em multivariate} direct quantization which preserves
feature information for each dimension with the manifold embedding representations (MVDFQ/MVDFQM)
of the original multivariate (continuous-valued) sequences.

In the following, we first describe the direct feature quantization (DFQ)
representations (Sec.~\ref{Sec:dfq}) and contrast them with vector quantization / codebook
based representations. We then define a novel family of
kernels on these multivariate DFQ representations (Sec.~\ref{Sec:MVDFQSK}).

\subsection{Direct feature quantization}
\label{Sec:dfq}
A multivariate direct feature quantization (MVDFQ) representation of
the original continuous-valued
multivariate sequence
\[X=(x_1,x_2,\ldots,x_n),x_i\in {\cal R}^R\]
is obtained by the direct quantization of each of the $R$ feature dimensions.
In this approach, each $j$th feature $f^j, j=1\ldots R$ is quantized by
dividing its range $(f^j_{min},f^j_{max})$ into the
finite number of intervals, $B$.

In the simplest case, the intervals can be defined, for instance,
using a uniform quantization with a pre-specified number of bins $B$, where the entire feature data range 
is divided into $B$ equal intervals of length $\delta=(f_{max}-f_{min})/B$
and the index of the quantized feature value $Q(f) = \lfloor (f - f_{min})/\delta \rfloor$ 
is used to represent the feature value $f$.

Partitioning of the feature data range could also be obtained
by using 1D clustering, e.g., $k$-means, to adaptively choose
dicretization levels and the number of bins for each dimension.
Discretization levels also can be chosen using, for example,
 Gaussian distribution assumption (see, e.g., \cite{SAX2003}) as
breakpoints under Gaussian curve producing equal-sized areas.

Varying the number of quantization levels $B$ will result in
more accurate (larger $B$) or more coarse (smaller $B$)
representation of the original real-valued data.
Here we choose appropriate number of quantization levels $B$
using a small scale cross-validation experiments on the subset of the
training data.

%For normalized data (e.g., $\[0,1\]$ range)

\begin{figure*}[!th]
\centering
\includegraphics[width=0.7\textwidth]{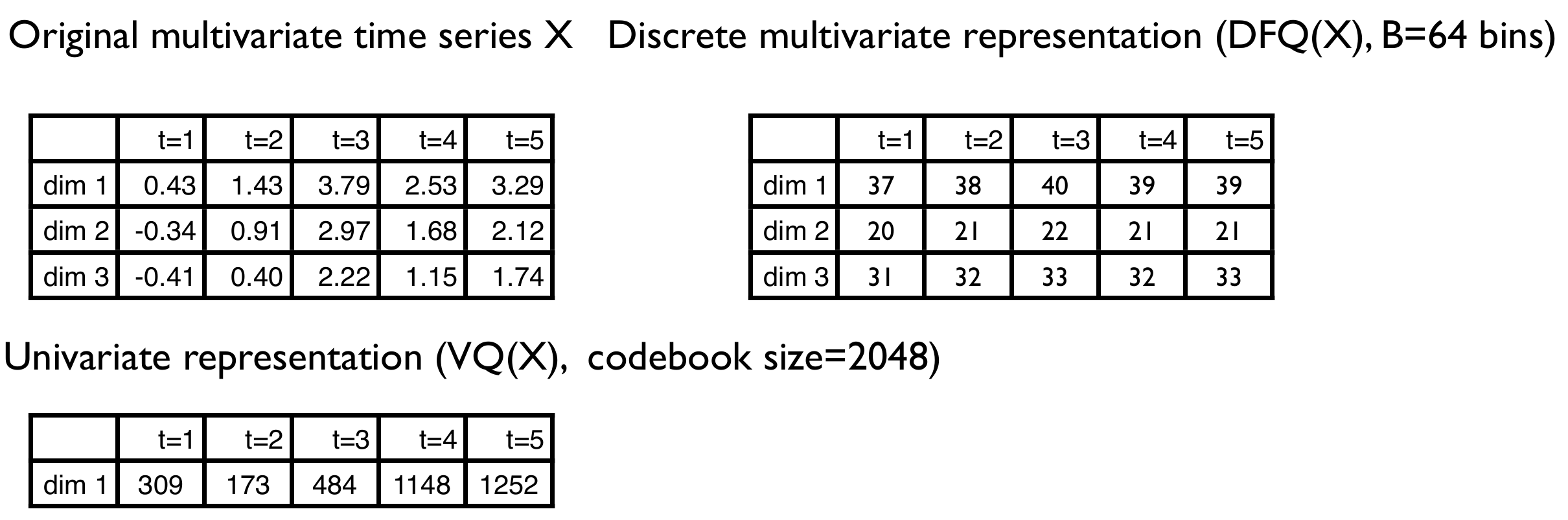}
\caption{The proposed discrete multivariate representation (DFQ). The discrete representation DFQ(X) is obtained from the original $R$=3-dimensional continuous-valued multivariate sequence $X$ by directly discretizing each of $R$ feature dimension. In contrast, the typically used vector quantization approach (codebook) represents
multivariate sequence as a one-dimensional discrete sequence of codeword indices.}
\label{Fig:dfq}
\end{figure*}

Figure~\ref{Fig:dfq} shows an example of a DFQ representation for
the 3-dimensional time series $X$ ($R$=3) where the 3-dimensional
DFQ representation has been obtained using a uniform binning ($B$=64)
along each of the three data dimensions.
As can be seen from the figure, compared to the vector quantization
approach, the DFQ retains feature values along each dimension, thus
providing a more accurate description of the original real-valued
sequence.

We will show in the experiments that using DFQ 
{\em multivariate representations} and MVDFQ kernels described below
can {\em significantly (by 25-40\%) improve} predictive accuracy compared
to traditional
1D (univariate) kernel representations as well as other state-of-the-art approaches
(Sec.~\ref{Sec:exp}).

\subsection{Multivariate Direct Feature Quantization Similarity Kernels}
\label{Sec:MVDFQSK}
In the following, we first define an efficient multivariate DFQ similarity
kernel (MVDFQ-SK) $K(DFQ(X),DFQ(X))$ for the direct feature quantization
(DFQ) representation defined in Sec.~\ref{Sec:dfq}.
We then present MVDFQ with the manifold embedding (MVDFQM) in Sec.~\ref{Sec:manifold} that as we show experimentally further improve
predictive ability of the classifiers on a number of challenging tasks and
datasets.
%that can be evaluated in {\em linear time}.

To compute similarity between two multivariate sequences $X$ and $Y$,
we propose a kernel function defined as

%Similar to the kernel proposed in~\cite{sdm2012distsk} for comparing sequences under {\em binary Hamming} embedding, we define the similarity kernel between two multivariate sequences $X$ and $Y$ under DFQ representation (Sec~\ref{Sec:dfq}) as
\begin{eqnarray}
 & K_{MVDFQ}(DFQ(X),DFQ(Y)) & = \nonumber\\
 & \displaystyle\sum_{\alpha_{R \times k} \in DFQ(X)} \sum_{\beta_{R \times k} \in DFQ(Y)} {\cal K}(\alpha_{R \times k},\beta_{R \times k})
\label{Eq:2Dkernel}
\end{eqnarray}
where $\alpha_{R \times k}$ and $\beta_{R \times k}$ are $R \times k$ submatrices contained in $DFQ(X)$ and $DFQ(Y)$
and ${\cal K}(\alpha_{R \times k}, \beta_{R \times k})$ is a kernel
function defined for measuring similarity between two $R\times k$ submatrices. 
Similarly to $k$-mer based kernel functions (e.g., Eq.~\ref{Eq:MK}), this kernel function essentially computes the similarity between sequences by
a cumulative comparison of all pairs of $R \times k$ submatrices contained in $DFQ(X)$ and $DFQ(Y)$ using a submarix kernel function ${\cal K(\cdot,\cdot)}$.

One natural definition for the submatrix kernel ${\cal K}(\cdot,\cdot)$  is cumulative {\em row-based comparison}
\begin{equation}
{\cal K}(\alpha_{R \times k},\beta_{R \times k}) = \sum_{r=1}^{R} I_{1 \times k}(\alpha_{R \times k}^r,\beta_{R \times k}^{r})
\label{Eq:2Dsubkernel}
\end{equation}
where $I_{1 \times k}(\cdot,\cdot)$ is a similarity/indicator function for
matching 1D rows $\alpha_{R \times k}^r$ and $\beta_{R \times k}^{r}$.
The matching function $I_{1 \times k}(\cdot,\cdot)$ could be defined
as $I_{1 \times k}(\alpha,\beta)=1$ if $d(\alpha,\beta)\leq m$, and 0 otherwise (similar to
the mismatch kernel).

In the experiments, we use the state-of-the-art spectrum, mismatch~\cite{MISMATCH-K}, and
spatial sample (SSSK)~\cite{icpr2010sssk}
kernel functions as our one-dimensional row matching function
\[I_{1\times k}(\alpha_{R\times k}^r,\beta_{R\times k}^r)\]
in Eq.~\ref{Eq:2Dsubkernel}, which results
in corresponding {\em multivariate} DFQ spectrum, mismatch, and spatial
sample kernels (referred as MVDFQ-Spectrum, MVDFQ-Mismatch, and MVDFQ-SSSK, respectively).

Intuitively, according to the kernel definition (Eq.~\ref{Eq:2Dsubkernel}),
similar $R\times k$ submatrices (i.e. submatrices with many similar rows)
will result in a large kernel value ${\cal K}(\cdot,\cdot)$.

Using Eq.~\ref{Eq:2Dsubkernel}, the multivariate DFQ kernel in Eq.~\ref{Eq:2Dkernel} can be
written as 
\begin{eqnarray}
& K_{MVDFQ}(X,Y)  = \nonumber\\
 &\displaystyle\sum_{r=1}^{R} \sum_{\alpha_{R \times k} \in DFQ(X)} \sum_{\beta_{R \times k} \in DFQ(Y)}
I_{1\times k}(\alpha_{R \times k}^{r},\beta_{R \times k}^{r}) 
\label{Eq:MVDFQkernel}
\end{eqnarray}
which can be efficiently computed by running the corresponding kernel
with a 1D $k$-mer matching function $I_{1 \times k}(\cdot,\cdot)$ $B$ times, i.e. for each row $b=1\ldots R$.
The overall complexity of evaluating multivariate kernel $K_{MVDFQ}(DFQ(X),DFQ(Y))$ for
two $R$-dim DFQ sequences $DFQ(X)$ and $DFQ(Y)$ is then $O(R\cdot k \cdot n)$, i.e. is {\em linear} in the sequence length $n=|X|$ and the number
of dimensions $R$.

\subsection{Manifold embedding}
\label{Sec:manifold}
While typical string kernel methods assume Euclidean feature space
and use Euclidean distance,
a {\em probabilistic manifold assumption} on the geometry of the data space
could be more natural and effective (see e.g.~\cite{multinomial2005text,jebara2004ppk}).
Given $d$-dim feature representation of a sequence,
$\Phi(X)=(\phi_1(X),\ldots,\phi_d(X))$,
the sequence $X$ can be considered as a point on the multinomial manifold using
L1 embedding of $\Phi(X)$:
\begin{equation}
  \hat{\Phi}(X) = \left(\frac{\phi_1(X)}{\sum_i \phi_i(X)}, \ldots,\frac{\phi_d(X)}{\sum_i \phi_i(X)}\right)
\end{equation}
where, e.g., in the simple $k$-mer frequency representation,
$\phi_i(X)=f(k_i,X)$, the frequency of $k$-mer  $k_i$ in sequence X.

Then, a natural measure of affinity between the distributions $\hat{\Phi}(X)$ and $\hat{\Phi}(Y)$ on 
the multinomial manifold is a Bhattacharyya affinity~\cite{bhatt1943}, i.e.

%For example, under a {\em multinomial manifold} assumption, the kernel/distance
%$K_{manifold}(X,Y)$ between $\Phi(X)$ and $\Phi(Y)$ could correspond to
%Bhattacharyya affinity measure between two probability distributions , i.e.
\begin{align}
\nonumber K_{manifold}(X,Y) & = <\displaystyle\sqrt{\hat{\Phi}(X)},\sqrt{\hat{\Phi}(Y)}> \\
\nonumber & = \sum_i \sqrt{\hat{\phi}_i(X)}\sqrt{\hat{\phi}_i(Y)}
\end{align} 

Using the equation above and equations for MVDFQ Eq.~\ref{Eq:MVDFQkernel},
we obtain the MVDFQ kernel with the {\em manifold} embedding (MVDFQM):
\begin{equation}
K_{MVDFQM}(X,Y) = \sum_{r=1}^{R} \sum_{\gamma} \sqrt{\phi^{r}_{\gamma}(X)}\sqrt{\phi^{r}_{\gamma}(Y)} 
\end{equation} 
where $\gamma \in \{1,\ldots,B\}^K$ is a $k$-mer over the discretization
alphabet $\Sigma = \{1,\ldots,B\}$ and 
\begin{equation}
\phi^{r}_{\gamma}(X) = \sum_{\alpha^{r}_{R \times k} \in DFQ(X) } I(\alpha_{R\times k}^{r},\gamma)
\end{equation}
is the
number of occurrences of $\gamma$ in the $r$-th dimension/row of $X$.

In the experiments, we test the manifold embedding with MVDFQ kernel as
well as other standard string kernels (we will refer to the MVDFQ with
the manifold embedding as MVDFQM, and to the standard VQ kernels with
the manifold embedding as VQ-M).
%We will demonstrate the utility of the proposed MVDFQM kernels and
%the manifold embedding in enhancing classification performance.

\subsection{Advantages of multivariate DFQ}
The proposed multivariate DFQ kernel method has the following merits:
\begin{itemize}
\item
It improves the predictive ability of typical discrete univariate kernel
methods with VQ by applying them jointly to multiple discrete sequences obtained from direct discretization of each data dimension of the original real-valued
multidimensional sequence.
\item
Unlike the state-of-the-art approach of quantizing high-dimensional data
samples into codewords,
it allows for classifier to learn importance of
each feature for classification,
as the significance of each data dimension for classification can
be different.

\item
It does not rely on clustering or binary similarity-preserving
hashing techniques (e.g., as in~\cite{sdm2012distsk}) as it directly discretizes the feature space using, e.g., uniform binning or adaptive
clustering algorithm ($k$-means).
\item
It has a low computational cost as it runs in linear time and is scalable to large sequence data sets.
\item
It can be used with any of the existing univariate sequence kernels (mismatch/spectrum~\cite{MISMATCH-K}, kernels~\cite{icpr2010sssk}, gapped/subsequence kernels~\cite{LodhiString,fastInexact}, etc) to improve performance.
\end{itemize}

\section{Experimental evaluation}
\label{Sec:exp}
We study the performance of our methods in terms of the predictive
accuracy and the running time on a number of challenging
sequence classification problems using standard benchmark datasets
for the music genre classification and artist recognition, as well as
protein sequence analysis.

\begin{table*}[!ht]
\centering
\caption{Benchmark datasets}
\label{Tab:datasets}
%\small{
\begin{tabular}{lccl}
\hline
Dataset & \# seqs & \# classes & Evaluation\\
\hline\hline
Music genre~\cite{tzan2002,sigir2003music} & 1000 & 10 & 5-fold cross-validation error\\
ISMIR2004 contest & 1458 & 6 & 5-fold cross-validation error\\
Artist~\cite{timbreellis2007music} & 1413 & 20 & 6-fold hold-one-album-per-artist validation error\\
SCOP protein homology~\cite{JasonWeston08012005} & 7329 & 54 & 54 binary remote homology detection tasks\\
\hline
\end{tabular}
%}
\end{table*}

\begin{table*}[!hbt]
\centering
\caption{Music genre recognition (10-class). Comparison
of the error rates (\%) for the vector quantization (VQ), similarity hashing~\cite{sdm2012distsk},
and the proposed multivariate direct feature quantization (MVDFQ).
All methods are compared with (the right three columns) and without
(three columns on the left) the proposed manifold embedding (Sec.~\ref{Sec:manifold}).
MVDFQ performs better than traditional VQ kernels and sim. hashing
kernels~\cite{sdm2012distsk}}
\label{Tab:genre}
%\small{
\begin{tabular}{lccc||ccc}
\hline
Method & VQ & Sim. hashing~\cite{sdm2012distsk} & MVDFQ
& VQ-M & Sim. hashing-M & MVDFQM\\
\hline\hline
% TODO: 1. add variances, 2.mismatch DFQ, 3.sim. hashing SSSK
Spectrum & 34.5 & 29.2 & 28.5 & 26.9 & 24.7 & 23.0 \\
Mismatch & 32.6 & 29.7 & 24.7 & 31.2 & 23.4 & 20.3 \\
SSSK     & 31.1 & 25.9 & {\bf 22.6} & 25.0 & 23.0 & {\bf 17.3} \\
\hline
\end{tabular}
%}
\end{table*}

\subsection{Datasets and experimental setup}
We test proposed methods on a number of multi-class sequence classification tasks:
\begin{enumerate}
\itemsep 0pt
\item
10-class music genre classification\footnote{http://opihi.cs.uvic.ca/sound/genres}. This dataset is a reference music genre recognition dataset
introduced in~\cite{tzan2002,sigir2003music}. It contains 1000 30-sec song
fragments grouped into 10 genres (blues, rock, classical, etc.), with
each genre represented by 100 songs. The task
here is to correctly predict the genre of the musical sample.
\item
6-class music genre recognition (ISMIR contest\footnote{http://ismir2004.ismir.net/genre\_contest/index.htm}). This is a benchmark music genre recognition task with samples classified into 6 genres.
\item
20-class music artist identification (artist20 dataset\footnote{http://labrosa.ee.columbia.edu/projects/artistid/}). This benchmark dataset
contains songs from 120 albums (6 albums per artist) with the
task of correctly identifying artist for songs from previously unseen
albums.
\item
protein remote homology detection 
(7329 sequences, 54 experiments)~\cite{JasonWeston08012005,JMLRProfileMultiClass}.
The task here is to correctly infer membership of a given protein
in protein superfamilies. % with 54 superfamilies as targets.
\end{enumerate}
Table~\ref{Tab:datasets} provides details of the datasets used in the
experiments.

For all music classification tasks input sequences are multivariate sequences of 13-dimensional MFCC feature vectors.

\subsection{Baseline methods}
We compare the proposed multivariate direct feature quantization (MVDFQ) kernel approach
for the multivariate sequence classification to three related baselines:
\begin{enumerate}
\itemsep 0pt
\item
Traditional vector quantization (VQ) approaches with univariate
string kernels. This approach has been used in a number of previous studies
for the image categorization, text analysis, music classification (see e.g., ~\cite{smkcvpr,sdm2010wcl,icpr2010sssk}).
\item
A similarity hashing-based kernel approach described recently in~\cite{sdm2012distsk} (a Euclidean similarity-preserving binary Hamming embedding of original real-valued MFCC feature vectors).
\item
A multivariate VQ approach using multiple codebooks of different sizes.
In this approach, VQ representations are stacked to obtain essentially a multivariate VQ codebook sequence representation.
\end{enumerate}

We also compare with a number of other state-of-the-art methods specifically developed for
the music sequence classification, namely
multivariate autoregressive models~\cite{autoregmfcc2010},
multilinear models~\cite{multilin2008music},
as well as methods with more problem-specific and sophisticated
features (aggregate Adaboost~\cite{aggregateAdaboost}, classifier fusion with rich spectral and cepstral features~\cite{spectralcepstral2009}, non-negative matrix factorization-based approaches~\cite{nmf2008music}).

We test our methods using the state-of-the-art spectrum/mismatch~\cite{fastInexact} and spatial (SSSK)~\cite{icpr2010sssk} kernels as our
basic univariate kernels, i.e. to implement row matching functions $I_{1\times k}(\cdot,\cdot)$ in
Eq.~\ref{Eq:2Dkernel}).

%(i.e. we use spectrum, mismatch, and spatial sample kernels (SSSK) as
%row matching function $I_{1\times k}(\cdot,\cdot)$ in
%Eq.~\ref{Eq:2Dkernel}).

We use Support Vector Machines (SVMs) classifiers with all kernels.

We also explore the performance impact of varying
the number of codewords, discretization bins,
a discretization algorithm (uniform, $k$-means).
% as well as a choice of
%the baseline univariate string kernel matching functions.

\subsection{Evaluation measures}
For music genre classification experiments a standard 5-fold cross-validation procedure is used as in previous studies~\cite{sigir2003music,aggregateAdaboost} to evaluate classification performance.
For music artist recognition,
we follow a 6-fold leave-one-album-out validation procedure proposed in
the previous work~\cite{timbreellis2007music}.

We report average multi-class classification errors as well as F1 scores for all tasks.

\subsection{Parameters and settings}
For the vector quantization (VQ) models,
we construct codebooks with 2048 codewords from input MFCC vectors.
For the multiple-codebook VQ method, we use codebooks with 1024, 2048, an 4096 codewords.

We test our direct feature quantization approach using
(1) a uniform quantization of each feature dimension into a fixed number
of bins ($B$=32) and (2)
using $k$-means clustering along each dimension to adaptively select
quantization levels and the number of bins per dimension.
The number of bins have been found from an initial cross-validation
on the subset of training data.

During the testing (classification), for input values outside
of the $(f_{min},f_{max})$ range, we use special values of 0 and $B+1$ for
values smaller than $f_{min}$ or larger than $f_{max}$.

For the discrete embedding with similarity hashing~\cite{sdm2012distsk},
we set the number of bits $E=32$ which has been found to perform
well in~\cite{sdm2012distsk}.

The length of $k$-mers used in the MVDFQ-spectrum/mismatch kernels
has been set to $k$=6 and and the number of mismatches is set as $m$=1
for the mismatch kernels (these values have been selected using
cross-validation on the subset of the training data).

For the univariate kernels on codeword (VQ) settings,
the best settings of the $k$-mer length and the
number of mismatches are similar, $k=5-6$ and $m$=1,2.

For the spatial sample kernels (SSSK)~\cite{icpr2010sssk}
we use the spatial kernel with three ($t$=3) $k$=1-mer features and
the maximum distance parameter $d$=5.

All experiments are performed on a single $2.8GHz$ CPU.
The datasets used in our experiments and the supplementary data/code are
available at\\
\texttt{http://pkuksa.org/\textasciitilde pkuksa/mvdfq.html}.

%parameters
% spectrum k=1, mismatch(5,2)
% triple k=2,t=3,d=5
% sim hashing k=8

\begin{table*}[!ht]
\centering
\caption{Music genre recognition (10-class). Comparison with previous
results and baselines.}%, 13-dim. MFCC features only)}
\label{Tab:genre_comp}
%\small{
\begin{tabular}{lcc}
\hline
{\bf method} & Error, \% & F1 \\
\hline\hline
Baseline 1: MFCC (Mean+Variance) & 48.30 & 51.55 \\
Baseline 2: NMF~\cite{nmf2008music} & 26.0 & -\\
Baseline 3: Multiple codebook VQ & 29.6 & 70.51 \\
Baseline 3: (non-MFCC): DWCH~\cite{sigir2003music} & 21.5 & - \\
Baseline 4: MAR (multivariate autoregressive model)~\cite{autoregmfcc2010} & 21.7 & -\\
Baseline 5: AdaBoost (MFCC,FFT,LPC,etc)~\cite{aggregateAdaboost} & 17.5 & - \\
%Baseline 6: Classifier fusion (MFCC,NASE,OSC)~\cite{spectralcepstral2009} & 9.4 & -\\
%Baseline 6: DBN~\cite{dbn2010hamel} & 15.7 & - \\
\hline
MVDFQ-SSSK (MFCC) & 26.3 & 73.88 \\
MVDFQM-SSSK (MFCC) & 17.3 & 82.85\\%{\bf 12.3} & {\bf 88.19} \\ 
MVDFQM-SSSK (MFCC+FFT64) & {\bf 13.6} & {\bf 86.53}\\%{\bf 9.3} & {\bf 91.05} \\ 
\hline
\end{tabular}
%}
\end{table*}

\if 0
\begin{table*}[!ht]
%\tabcolsep 1pt
\centering
\caption{Music genre recognition (10-class)}%, 13-dim. MFCC features only)}
\label{Tab:genre}
\begin{tabular}{lcc}
\hline
{\bf method} & Error, \% & F1 \\
\hline\hline
Baseline 1: MFCC (Mean+Variance) & 48.30 & 51.55 \\
Baseline 2: NMF~\cite{nmf2008music} & 26.0 & -\\
Baseline 3: (non-MFCC): DWCH~\cite{sigir2003music} & 21.5 & - \\
Baseline 4: MAR (multivariate autoregressive model)~\cite{autoregmfcc2010} & 21.7 & -\\
Baseline 5: AdaBoost (MFCC,FFT,LPC,etc)~\cite{aggregateAdaboost} & 17.5 & - \\
Baseline 6: Classifier fusion (MFCC,NASE,OSC)~\cite{spectralcepstral2009} & 9.4 & -\\
%Baseline 6: DBN~\cite{dbn2010hamel} & 15.7 & - \\
\hline
{\bf VQ vs Sim. hashing~\cite{sdm2012distsk} vs DFQ}\\
\hline
VQ Spectrum & 34.5 & 65.61 \\
VQ Mismatch & 32.6 & 67.51 \\
VQ SSSK & 31.1 & 69.08 \\
Multiple VQ Spectrum & 29.6 & 70.51 \\
Sim. hashing~\cite{sdm2012distsk} Spectrum & 27.8 & 72.25 \\ %k=8,B=64
MVDFQ-Spectrum & 28.5 & 72.06 \\
MVDFQ-SSSK & {\bf 26.3} & {\bf 73.88} \\
\hline
{\bf VQ vs Sim. Hashing~\cite{sdm2012distsk} vs DFQ (Manifold embedding)}\\
\hline
VQ Manifold Spectrum & 26.9 & 73.31 \\
VQ Manifold SSSK & 25.0 & 75.27 \\ 
Sim. hashing~\cite{sdm2012distsk} Manifold Spectrum & 24.1 & 76.29 \\
MVDFQ-1D Manifold SSSK & 22.0 & 78.07 \\
MVDFQ Manifold Spectrum & 23.0 & 77.23 \\
MVDFQ Manifold Spectrum ($k$-means) & 22.8 & 77.41 \\
MVDFQ Manifold SSSK & 17.3 & 82.85\\%{\bf 12.3} & {\bf 88.19} \\ 
MVDFQ Manifold SSSK+FFT64 & 13.6 & 86.53\\%{\bf 9.3} & {\bf 91.05} \\ 
\hline
\end{tabular}
\end{table*}
\fi

\subsection{Music genre recognition} 

We first compare the proposed multivariate DFQ kernel approaches (MVDFQ)
(Sec.~\ref{Sec:MVDFQSK},~\ref{Sec:manifold})
with the vector quantization-based univariate kernels, and
the recently proposed binary Hamming similarity hashing kernels~\cite{sdm2012distsk}. 

All of the methods are compared with and without the proposed
manifold embedding (Sec.~\ref{Sec:manifold}).

As shown in Table~\ref{Tab:genre}, on a widely used benchmark dataset for
music genre recognition~\cite{sigir2003music,tzan2002}
(10 genres, each with 100 sequences),
proposed {\em multivariate} DFQ kernels
improve over traditional univariate VQ kernels, as well as
recently proposed binary Hamming similarity hashing
kernels~\cite{sdm2012distsk}.
Using direct feature quantization (DFQ) kernels
effectively improves accuracy compared to the VQ, and the similarity hashing
approach for all basic kernels (spectrum, mismatch, and sparse spatial
sample kernels (SSSK)).
For instance, MVDFQ-SSSK achieves a significantly
lower error rate of 22.6\% compared to 31.1\% using VQ or 25.9\% using
similarity hashing (27\% and 13\% improvements, respectively).

Using the $k$-means clustering for the discretization with DFQ results
in the performance similar to the uniform quantization (e.g.,
MVDFQ-Spectrum and MVDFQ-SSSK with $k$-means achieve slightly lower
errors of 22.8\% and 16.9\% compared to 23.0\% and 17.2\% with
the uniform quantization).

We also note that using the manifold embedding further reduces error
for all of the methods, including the VQ-based and similarity
hashing kernels. These consistent improvements in accuracy across all of the
methods, could be attributed to the manifold embedding
effectively exploiting intrinsic geometric structure of music data.
Overall, MVDFQ with manifold embedding (MVDFQM-SSSK) achieves the best error rate of 17\% compared
to the the best 23\% error rate of similarity hashing kernel or 
25.0\% using VQ. (24\% and 30\% relative improvements in the
error rates, respectively).

We also compare with previous best results on this music genre
recognition dataset and baselines
in Table~\ref{Tab:genre_comp},  including multivariate autoregressive
models~\cite{autoregmfcc2010}, wavelet-based DWCH~\cite{sigir2003music} method,
aggregate AdaBoost~\cite{aggregateAdaboost},
approaches specifically developed for
the music classification that also use many other features in addition to MFCC.
As can be seen from the table, using multivariate DFQ kernels (MVDFQ-SK)
compares well with the state-of-the-art results (e.g., AdaBoost
method~\cite{aggregateAdaboost} with much richer feature set). 

The proposed MVDFQM method is also more effective than using
the multiple codebook VQ method.
%achieves significantly higher accuracy of 87.7\% compared to only 75.0\%
%when using best univariate kernels (a 50\% relative improvement in error rate).
We also note that expanding feature set by adding a set of 64 FFT features
to the 13 MFCC feature set (i.e. $77\times|X|$ multivariate representation),
could further increase accuracy to 86.4\% compared to that of 82.7\% with MFCC features alone (Table~\ref{Tab:genre_comp}).

We also note the utility of the {\em multivariate} direct feature
quantization representation and kernel (MVDFQ) as opposed to
the {\em univariate}  representations: the univariate kernel on the one-dimensional (1D) sequence
obtained from the DFQ multivariate sequence by encoding each $R$=13-dim feature vector as one codeword $c=\sum_i^R Q(f_i) B^{i-1}$, i.e. using alphabet size $|\Sigma|=B^R$,
gives a higher error rate of 22\% compared to that of 17.3\% when using the {\em multivariate} DFQ kernel with the manifold embedding (MVDFQM).

\begin{table*}[!ht]
\tabcolsep 1pt
%\begin{minipage}[t]{0.5\linewidth}
\centering
\caption{Comparison of the vector quantization (VQ), the similarity hashing~\cite{sdm2012distsk} approach, and the proposed multivariate DFQ method (MVDFQ) on the music genre recognition ISMIR contest, error rates (\%)}
% 13-dim. MFCC features only)}
\label{Tab:ismir}
%\small{
\begin{tabular}{l|c||c|c}
\hline

{\bf method} & Error, \% & {\bf method} (with Manifold) &  Error, \% \\
\hline\hline
VQ & 24.15 & VQ-M & 19.62\\ %& 24.15 & 68.99 \\ %k=1
Sim. hashing~\cite{sdm2012distsk} & 22.63 & Sim. hashing-M &17.83\\ %& 22.63 & 71.57 \\ %k=6,B=32
MVDFQ & {\bf 19.89} & MVDFQM & {\bf 16.74}\\ %{\bf 16.74} & {\bf 80.57} \\ %k=3
%MVDFQ Manifold Spectrum+FFT64 & {\bf 16.19} & {\bf 80.79} \\ %k=3
\hline
\end{tabular}
%}
%\end{minipage}
\end{table*}
\begin{table*}[!ht]
%\tabcolsep 1pt
%\hfill
%\begin{minipage}[t]{.45\linewidth}
\centering
\caption{Classification performance on the ISMIR contest data}
%Comparison of univariate and multivariate kernels and previous results.}
% 13-dim. MFCC features only)}
\label{Tab:ismir_comp}
%\small{
\begin{tabular}{lcc}
\hline
{\bf method} & Error, \% & F1 \\
\hline\hline
Baseline 1: MFCC (Mean+Variance) & 34.85 & 58.15 \\
Baseline 2: Multilinear (Cortical)~\cite{multilin2008music} & 19.05 & - \\
Baseline 3: pLSA~\cite{plsa2009music} & 19.9 & - \\
Baseline 4 (best): NMF~\cite{nmf2008music} & 16.5 & - \\ 
MVDFQM-Spectrum (MFCC) & {\bf 16.74} & {\bf 80.57} \\ %k=3
MVDFQM-Spectrum (MFFC+FFT64) & {\bf 16.19} & {\bf 80.79} \\ %k=3
\hline
\end{tabular}
%}
%\end{minipage}
\end{table*}

\begin{table*}[!ht]
\centering
\caption{Music artist identification (20-class, 13-dim MFCC features only)}
\label{Tab:artistid}
%\small{
\begin{tabular}{lcc}
\hline
{\bf method} & Error, \% & F1 \\
\hline\hline
Baseline 1: MFCC (Mean+Variance) & 59.54 & 40.91 \\
Baseline 2: MFCC+FFT64 (Mean+Variance) & 54.17 & 47.09\\ 
Baseline 3: MFCC (GMM)~\cite{timbreellis2007music} & 44.0 & - \\
VQ & 42.97 & 57.26 \\ %k=1
Sim. hashing~\cite{sdm2012distsk} & 34.62 & 66.22 \\ %k=6
%Manifold Spectrum-($k$=1), $B$=64 & 28.61 & 71.76 \\
MVDFQM-Spectrum & {\bf 25.67} & {\bf 74.79} \\ %k=6
\hline
\end{tabular}
%}
\end{table*}

We observe similar overall improvements for multivariate MVDFQ-SK kernels
on another benchmark dataset (ISMIR2004 genre contest),
Table~\ref{Tab:ismir}.
For instance, using the MVDFQ-SK string kernel with uniform direct feature quantization (MFCC only)
reduces the error rate to 16.7\% compared to that of 19.6\% when using VQ
with the univariate kernel.

As also can be seen from the Table~\ref{Tab:ismir_comp}, obtained error rates (16-17\%) for ISMIR genre recognition compare well with a number
of previous best results, including the recent non-negative matrix
factorization method~\cite{nmf2008music}(16.5\% error), or complex auditory model
and cortical representation of~\cite{multilin2008music} (19\% error).
%Music genre recognition is a particularly interesting problem
%in our setting because original representations of music data are typically that of sequences of continuous-valued feature vectors (e.g., 13-dim. MFCC features).
%and the DWCHs method~\cite{sigir2003music}, an approach specifically 
%developed for music classification.  
%Using the raw 13-dim. MFCC features with direct uniform feature
%quantization achieve $19.2$ error rate. 

\subsection{Artist recognition}
We also illustrate the utility of our multivariate MVDFQ-SK kernels and representations on the multi-class artist identification problem
using the standard {\em artist20} dataset
with 20 artists, 6 albums each (1413 tracks total).
Table~\ref{Tab:artistid} lists results for the 6-fold album-wise cross-validation 
with one album per artist held out for testing.
Using multivariate MVDFQ-SK kernels with the direct uniform quantization of
MFCC features
yields a much lower 25.7\% error compared to the best error rate for
42.9\% for univariate kernels (a 40\% relative improvement in error).

%\if 0
\subsection{Protein remote homology detection}
In Table~\ref{Tab:homology},
we compare our proposed {\em multivariate} DFQ string kernel method (using 20-dim BLOSUM rows as feature vectors for individual amino acids, i.e. $20\times|X|$ multivariate sequences)
with a number of state-of-the-art kernel methods for the
remote homology detection
including spectrum/mismatch kernels~\cite{MISMATCH-K,SPECTRUM-K},
spatial sample kernels~\cite{icpr2010sssk}, similarity hashing kernels~\cite{sdm2012distsk}, %semi-supervised
%cluster kernel~\cite{JasonWeston08012005},
%as well as state-of-the-art profile kernel~\cite{PROFILE-K}.
as well as recently proposed spectrum-RBF and mismatch-RBF
methods~\cite{bmc2010phychem} which also incorporate {\em physico-chemical
descriptors}. %with traditional spectrum/mismatch kernels.
%We test both similarity hashing and
%direct feature quantization approaches with our discrete multivariate string kernels.

As can be seen from results in Table~\ref{Tab:homology}, multivariate DFQ
string kernel (MVDFQ) provides effective improvements over
other methods.
%standard univariate string kernel
%approaches (spectrum/ mismatch, spatial SSSK kernels), as well as recent similarity-preserving hashing kernels~\cite{sdm2012distsk},
%and amino acid kernels with physico-chemical descriptors~\cite{bmc2010phychem}.
For instance, using MVDFQ spectrum and mismatch kernels with BLOSUM substitution profiles
significantly improves average ROC50 scores
from 27.91 and 41.92 to 43.29 and 49.17, respectively
(relative improvements of 50\% and 17\%), compared to traditional univariate 
spectrum/mismatch approaches.

%Similar improvements are observed when using the spatial sample kernel (SSSK)
%(the average ROC50 increases from 50.12 using 1D amino acid sequences
%to 55.54 using the MVDFQ BLOSUM representation with the SSSK kernel).

\if 0
We also observe that the MVDFQ kernel provides substantial
improvements in the semi-supervised setting. 
%using {\em semi-supervised} 
%cluster kernel~\cite{JasonWeston08012005} and profile kernel approaches.
The multivariate DFQ kernel on sequence profiles used by the
profile kernel (obtained from the non-redundant sequence database (NRDB)~\cite{PROFILE-K}) achieves a higher average ROC50 score of 86.27 compared
to 81.51 of the profile kernel.
\fi

\begin{table*}[!ht]
\centering
\caption{Classification performance (mean ROC50) on protein remote homology detection (54 experiments). Multivariate DFQ string kernels perform better than
traditional (univariate) kernels and similarity-preserving hashing kernels~\cite{sdm2012distsk} ($p$-values indicate statistical significance of
the observed differences between univariate and corresponding multivariate DFQ kernels)}
\label{Tab:homology}
%\small{
\begin{tabular}{lcccc}
\hline
       & \multicolumn{2}{c}{Mean ROC50} & \\
Method & Univariate  & Sim. hashing~\cite{sdm2012distsk} & Multivariate DFQ & $p$-value\\
\hline\hline
Spectrum~\cite{SPECTRUM-K} & 27.91 & 35.14 & {\bf 43.29} & 2.5e-6\\
Mismatch-($k$=5,$m$=1)~\cite{MISMATCH-K} & 41.92 & 46.35 & {\bf 49.17} & 6.5e-3 \\
Spatial sample (SSSK)~\cite{icpr2010sssk} & 50.12 & 52.0 & {\bf 55.54} & 2.7e-5\\
%Profile kernel-($k$=5,$\sigma$=7.5) (NRDB)~\cite{PROFILE-K} & 81.51 & - & {\bf 86.27} & 2.9e-4\\
Spectrum-RBF~\cite{bmc2010phychem} & 42.1 & - & - & -\\
Mismatch-RBF~\cite{bmc2010phychem} & 43.6 & - & - & -\\
%Baseline 6: Semi-supervised Cluster kernel~\cite{JasonWeston08012005} & 67.91\\
%Spectrum (2D BLOSUM, sim. hashing) & 38.68\\
%Mismatch (2D BLOSUM, sim. hashing) & 44.05\\
%Spectrum (2D BLOSUM) & 43.29 & 2.5e-6\\
%Mismatch (2D BLOSUM) & 49.17 & 6.5e-3\\
%Spatial sample (SSSK) (2D BLOSUM)     & 55.54 & 2.7e-5\\
%Semi-supervised Cluster kernel (2D BLOSUM) & 70.14 \\
%Profile kernel (2D) & 86.27 & 2.9e-4\\
\hline
\end{tabular}
%}
\end{table*}

%\fi

\subsection{Running time} 
In Table~\ref{Tab:runTimeAnl}, we compare
the classification performance and the running time 
of our method to the recent binary Hamming embedding (similarity hashing)~\cite{sdm2012distsk}, and vector quantization-based univariate
kernels.
We vary the dimensionality of the embedding space $E$,
the codebook size, and the number of discretization bins $B$, respectively.
We note that for mismatch-($k$,$m$) kernel computation we use
the linear time sufficient-statistic based algorithm
from~\cite{nips2008inexact}) and for all other methods we use their
existing state-of-the-art implementations.

As can be seen from Table~\ref{Tab:runTimeAnl}, multivariate kernels
with the direct feature quantization (MVDFQ) display a better performance
compared to the similarity hashing~\cite{sdm2012distsk} and traditional univariate kernels.

\begin{table*}[!ht]
\centering
\caption{Music genre classification performance and running time for the kernels using the vector quantization, similarity hashing~\cite{sdm2012distsk}, and the proposed direct feature quantization (DFQ) approach
as a function of the codebook size, the embedding size, and the number of discretization bins, respectively} %Running times are given for
%$1000\times 1000$ kernel matrix computation.}
\label{Tab:runTimeAnl}
%\small{
\begin{tabular}{lcc}
\hline
Embedding size & Error, \% & \begin{mybox}Running time (s), $1000\times 1000$ kernel\\matrix computation\end{mybox}\\ 
\hline\hline
Mismatch(5,2) VQ $|\Sigma|$=2048 & 32.6 & 28 \\
Mismatch(5,2) VQ $|\Sigma|$=1024 & 32.5 & 26\\
Sim. hashing E=16,k=5 & 28.3 & 24 \\
Sim. hashing E=32,k=5 & 24.7 & 61 \\
Sim. hashing E=64,k=5 & 28.2 & 202 \\
MVDFQ B=16 & 23.9 & 9.4  \\
MVDFQ B=32 & 23.0 & 10.4 \\
MVDFQ B=64 & 22.8 & 11.6  \\
\hline
\end{tabular}
%}
\end{table*}

\section{Conclusions}
We presented novel discrete multivariate direct feature quantization
kernel methods (MVDFQ-SK and MVDFQ-SK with the manifold embedding) 
for data in the form of sequences of feature vectors (as in music MFCC sequences,
biological sequence profiles, or image sequences).
The proposed approach directly exploits original multivariate
feature sequences to improve sequence classification as opposed
to using univariate codeword sequences.
On three music classification tasks as well as protein sequence classification this shows significant 25-40\%
improvements compared to the traditional codebook learning and state-of-the-art
sequence classification methods.

\begin{spacing}{0.9}
\bibliography{arxiv2014dfq}
\bibliographystyle{abbrv}
\end{spacing}

\end{document}